\documentclass{article}

\usepackage{PRIMEarxiv}

\usepackage[utf8]{inputenc} 
\usepackage[T1]{fontenc}    
\usepackage{hyperref}       
\usepackage{url}            
\usepackage{booktabs}       
\usepackage{amsfonts}       
\usepackage{nicefrac}       
\usepackage{microtype}      
\usepackage{lipsum}
\usepackage[arabic,farsi,english]{babel}
\usepackage{fancyhdr}       
\usepackage{subfigure}
\usepackage{multirow}
\usepackage{float}
\usepackage{amsmath}
\usepackage{natbib}
\usepackage{graphicx}       
\graphicspath{{media/}}     

\title{Agentic Username Suggestion and Multimodal Gender Detection in Online Platforms: Introducing the PNGT-26K Dataset
}

\author{
 Farbod Bijary \\
  Computer Engineering Department\\
  Amirkabir University of Technology\\
  \texttt{farbod.bijary@aut.ac.ir} \\
   \And
 Mohsen Ebadpour \\
  Computer Engineering Department\\
  Amirkabir University of Technology\\
  \texttt{M.Ebadpour@aut.ac.ir} \\
  \And
 Amirhosein Tajbakhsh \\
 Iran University of Science \& Technology \\
  \texttt{am\_tajbakhsh@ind.iust.ac.ir}
}

\begin{document}
\maketitle

\begin{abstract}
Persian names present unique challenges for natural language processing applications, particularly in gender detection and digital identity creation, due to transliteration inconsistencies and cultural-specific naming patterns. Existing tools exhibit significant performance degradation on Persian names, while the scarcity of comprehensive datasets further compounds these limitations. To address these challenges, the present research introduces \textit{PNGT-26K}, a comprehensive dataset of Persian names, their commonly associated gender, and their English \textit{transliteration}\footnote{``To represent or spell in the characters of another alphabet.'' \textit{\href{https://www.merriam-webster.com/dictionary/transliteration}{Merriam-Webster.com Dictionary}}, Accessed 15 Aug. 2025.}, consisting of approximately 26,000 tuples. As a demonstration of how this resource can be utilized, we also introduce two frameworks, namely \textit{Open Gender Detection} and \textit{Nominalist}. Open Gender Detection is a production-grade, ready-to-use framework for using existing data from a user, such as profile photo and name, to give a probabilistic guess about the person's gender. Nominalist, the second framework introduced by this paper, utilizes agentic AI to help users choose a username for their social media accounts on any platform. It can be easily integrated into any website to provide a better user experience. The PNGT-26K dataset\footnote{\url{https://huggingface.co/datasets/farbodbij/persian-gender-by-name}}, Nominalist\footnote{Nominalist: \url{https://github.com/farbodbj/Nominalist}} and Open Gender Detection \footnote{Open Gender Detection: \url{https://github.com/farbodbj/Open-Gender-Detection}} frameworks are publicly available on Github.
\end{abstract}


\section{Introduction}
The digital landscape increasingly demands sophisticated natural language processing tools that can handle diverse linguistic and cultural contexts. Persian, spoken by over 110 million people worldwide, represents a significant yet underserved segment in NLP applications, particularly in the domain of personal name processing. The challenges surrounding Persian name analysis are multifaceted, encompassing issues of transliteration inconsistency, cultural specificity, and the scarcity of comprehensive public datasets.

Gender detection from names has emerged as a critical component in various applications, from demographic analysis to personalized user experiences. However, existing commercial tools and research frameworks exhibit significant performance degradation when applied to non-Western names, with accuracy dropping below 82\% for names from certain regions \cite{vanhelene2024inferring}. This disparity stems from training datasets that inadequately represent global naming conventions and fail to capture the morphological and semantic patterns unique to different linguistic traditions.

The Persian language presents additional complexity through its transliteration challenges. The absence of standardized mapping from the Persian alphabet to Latin characters creates a one-to-many problem, where a single Persian name can have multiple plausible English spellings \cite{babelstreet2022persian}. This variability significantly impacts information retrieval systems and NLP applications, as demonstrated by studies showing that most Iranian academics have their names recorded in multiple forms across databases \cite{kazerani2018examining}.

To address these challenges, we present three main contributions. First, we introduce \textit{PNGT-26K}, a comprehensive dataset of approximately 26,000 Persian names with their gender associations and English transliterations, systematically curated to address the transliteration bottleneck. Second, we develop the \textit{Open Gender Detection} framework, a multimodal system that combines name-based and image-based inference to provide probabilistic gender predictions. Finally, we present \textit{Nominalist}, an agentic AI framework for intelligent username generation that leverages cultural awareness and personalization to create appropriate digital identities.

\section{Related Work}
\label{sec:headings}

\subsection{Automated Gender Detection: Methods and Biases}
The inference of gender from names has evolved from simple dictionary lookups to complex deep learning based architectures. Early methods relied on lexicon-based systems, matching names against curated lists like those from the US Social Security Administration (SSA) \cite{ssa_baby_names, mullen2014gender}. These were often enhanced with probabilistic data to handle ambiguity in the case of names with multiple assigned genders. However, their primary limitation is the inability to classify names not present in their databases.

A significant advancement came with character-based machine learning and deep learning models, which can learn predictive patterns including suffixes, prefixes, or character n-grams from the string of characters in a name, which can be a strong determiner of gender for a name \cite{hu2021name}. An example of such patterns is the dominant usage of names ending in "a" ("Sophia", "Sara", etc.) for women rather than men in the English language \cite{hu2021name}. Deep learning architectures such as LSTMs, CNNs, and Transformers have demonstrated high accuracy by identifying gender-indicative morphological features like prefixes and suffixes \cite{hu2021name}. Studies on large datasets of American, Brazilian, and Bangladeshi names have reported accuracies exceeding 90\% \cite{hu2021name}.

Despite these technical advances, state-of-the-art tools, including commercial APIs like Genderize.io and Gender API, which use proprietary datasets and algorithms, exhibit a significant performance drop when applied to Asian names \cite{vanhelene2024inferring}. While achieving over 98\% accuracy on German names, their performance on names from East Asian countries can fall below 82\% \cite{vanhelene2024inferring}. This performance imbalance stems from training data that is not globally representative and fails to capture the diverse naming conventions across cultures. 

\subsection{Multimodal and Probabilistic Gender Detection}
To overcome the limitations of single-modality systems, research has increasingly turned to multimodal approaches, particularly in the context of social media, where users provide a variety of data points. Frameworks have been developed to infer gender by fusing information from a user's profile, including their name, profile picture, and textual descriptions \cite{perez2017learning}. In the approach used by Perez et al. at \cite{perez2017learning}, textual and visual cues from users' social media accounts, including Twitter and Instagram, were utilized for gender prediction. For each user, photos were processed using ImageNet \citep{5206848} embeddings, and related textual information was converted to feature vectors using LDA \cite{blei2003latent}, LIWC\cite{pennebaker2001linguistic}, and heuristic-based methods. These feature vectors were fused and used for training a deep neural network, achieving an accuracy of more than 91\% on test sets. These systems often yield a probabilistic output rather than a deterministic one. Instead of simply classifying a user as male or female, they provide a probability score that reflects the model's confidence in its prediction. This aligns with commercial tools that return a confidence score alongside the gender prediction, acknowledging the inherent ambiguity in the task \cite{genderize_api}. 

\subsection{The Persian Language Context: Transliteration and Data Scarcity}
The application of NLP techniques to Persian names is complicated by two core issues: the lack of a standardized transliteration system and the scarcity of comprehensive public datasets. The absence of a one-to-one mapping from the Persian alphabet to the Latin alphabet creates a "one-to-many" problem, where a single Persian name can have numerous plausible English spellings \cite{babelstreet2022persian}. This is not a challenge unique to Persian; similar issues are well-documented for other languages using non-Latin scripts. For instance, Arabic names face significant ambiguity due to the Abjad nature of the script, which often omits short vowels \cite{abdulmageed2020arbert}, while languages using the Cyrillic alphabet, such as Russian, must contend with multiple competing transliteration standards leading to variant spellings. This inconsistency, rooted in phonological and orthographic differences, severely degrades the performance of information retrieval and NLP systems, as a single entity may be represented by multiple, distinct strings \cite{hassanzadeh2022effects, kazerani2018examining}. For example, studies have found that a majority of Iranian faculty members have their names recorded in two or more different forms in academic databases \cite{kazerani2018examining}. This transliteration bottleneck breaks the fundamental assumption of consistent identifiers that underpins most NLP models.

The problem is exacerbated by the limited number of publicly available datasets. While a comprehensive commercial resource, the CJKI Database of Persian Names, exists, its proprietary nature makes it inaccessible for open academic research \cite{cjki2023persian}. Publicly available datasets on platforms like Kaggle or bundled with open-source tools are often limited in size, lack systematic collection of transliterations \cite{titanz123_kaggle, zeoses_github, armanyazdi_github, peymanslh_npm}. 

\subsection{Username suggestion frameworks: Generating Digital Identities}
The creation of usernames, or digital anthroponyms, is a critical aspect of online identity. Research shows that users frequently derive usernames from their real names through predictable transformations like abbreviation or concatenation \cite{peris2016identifying}. Furthermore, the characteristics of a username, such as its length or pronounceability, can influence social perceptions like trustworthiness \cite{topolinski2017username}. This suggests that username generation should move beyond simple heuristics (e.g., appending numbers to a taken name) toward more sophisticated, personalized approaches.

Generative models are well-suited for this creative task. Character-level Recurrent Neural Networks (RNNs) and their variants like LSTMs can learn the underlying patterns of name formation to generate novel, plausible names and usernames \cite{sutskever2011generating, rahalkar2020name}. For transforming a real name into a username, sequence-to-sequence models are particularly effective, as they can learn complex, non-linear mapping rules from data \cite{sutskever2014sequence}.


\subsection{Synthesis and Research Gap}
This review highlights a tripartite research gap that our work addresses. First, existing gender detection tools are biased against non-Western names \cite{vanhelene2024inferring}, a problem exacerbated in Persian by a systemic transliteration bottleneck that fragments digital identity. The lack of a large-scale, public Persian name dataset that accounts for these transliteration variants is a major obstacle to developing more equitable NLP tools. Second, the conceptual link between real-world onomastics and the generation of digital usernames is underexplored. Finally, existing username suggestion systems are often simplistic and lack cultural awareness.

This paper directly addresses these gaps. The \textbf{PNGT-26K} dataset provides the first large, open, and systematically curated resource for Persian names with transliterations. The \textbf{Open Gender Detection} framework offers a practical application of this dataset. Finally, the \textbf{Nominalist} framework introduces a novel, agentic AI approach for generating culturally-aware and personalized digital identities, moving beyond the current state of the art.

\section{Methodology}
\label{sec:methods}

\subsection{The PNGT-26K Dataset: Curation and Structure}
\label{sec:dataset_creation}

\subsubsection{Data Sourcing}
\label{subsubsec:data_sourcing}
The data gathering process was started by exploring existing works from various sources, including Kaggle and Github, then, through simple random sampling, samples of 1000 entries in size were taken from each candidate and manually validated by native-speakers. Finally, 3 of the candidates were chosen for the compilation process: iranian-Names-Database-By-Gender\cite{iranianNamesDB}, persian-names\cite{persianNames}, persian-names-gender\cite{persianNamesGender}. Table 1 demonstrates the distribution of each used source.

\begin{table}[h]
 \caption{Distribution of male and female names}
  \centering
  \begin{tabular}{llll}
    \toprule
    Dataset name     & Male (\%) & Female (\%) & \#  \\
    \midrule
    iranian-Names-Database-By-Gender & 68 & 32 & 19882 \\
    persian-names & 51 & 49 & 8888 \\
    persian-names-gender & 56 & 44 & 8457 \\
    \bottomrule
  \end{tabular}
  \label{tab:table}
\end{table}

\subsubsection{Data Preprocessing and Normalization}
\label{subsubsec:data_preprocessing}
In Persian NLP use cases, involving textual data from digital media, character-level normalization is often required because certain characters have multiple Unicode representations. For example, the Persian letter "yeh" can be represented in two ways by a single character ⟨\FR{ی}⟩ (U+06CC) or as a combination of Arabic yeh ⟨\AR{ي}⟩ (U+064A) followed by a combined dot below (U+0652). Additionally, the Persian letter "kaf" can appear as a single character ⟨\FR{ک}⟩ (U+06A9) or as an Arabic kaf ⟨\AR{ك}⟩ (U+0643) with no visual distinction in some fonts\cite{doctor2024graphemicnormalizationpersoarabicscript}. Such characters could lead to unwanted duplications in the final results. To address this, the Hazm\cite{hazm} Python library was used for all data sources, as it is a commonly used library in Persian NLP tasks.

\subsubsection{Data compilation}
\label{subsubsec:data_compilation}
To unify different data formats of the initial preprocessed data, making them compatible with the intended format of the final result, all datasets were merged. First, the duplicate records that contained the same Persian name, English transliterations, and assigned gender were removed. To mitigate potential invalid or unreliable data, we employed a local instance of DeepSeek-R1-Distill-Qwen-32B, identified as one of the best-performing Large Language Models (LLMs) in various benchmarks \cite{deepseekai2025deepseekr1incentivizingreasoningcapability}. This LLM was used to label entries that may have been transliterated incorrectly or contained misspellings. Subsequently, these flagged records underwent manual review, allowing for necessary corrections or deletions to refine the dataset. This methodology aligns with previous studies that have demonstrated LLMs, such as ChatGPT-4, achieving near-human performance in name-based gender identification tasks \cite{rahman2024llms}. Table 2 summarizes treatments used for each of the initial datasets in the compilation process.

\begin{table}[h]
 \caption{Dataset-Action Summary}
  \centering
  \begin{tabular}{lccc}
    \toprule
    Action & iranian-Names-Database-By-Gender & persian-names & persian-names-gender \\
    \midrule
    Normalization & \checkmark & \checkmark & \checkmark \\
    Duplicate Removal & \checkmark & & \checkmark \\
    English Transliteration & & \checkmark & \\
    \bottomrule
  \end{tabular}
  \label{tab:dataset-actions}
\end{table}

\subsection{Open Gender Detection: A Multimodal Probabilistic Framework}
\label{sec:open_gender_detection}

\subsubsection{System Architecture}
\label{subsubsec:ogd_architecture}
The system is composed of two main modules. A module responsible for labeling photos of the user as male or female, by passing OpenCLIP \cite{radford2021learningtransferablevisualmodels} embeddings to an SVM trained on roughly 160K samples of profile pictures of men and women with equal distribution. The second module uses the name provided by the user in their profile. This name is searched in a name-gender dataset similar in formatting to PNGT-26K using a string distance function, top K names are retrieved, and used to determine a probability of the name belonging to a male or female person. Finally, the result of the two modules is placed in a vote box, and the fusion logic is applied to both results using a mediator function to return the final probabilistic guess.

\subsubsection{Name-Based Inference Model}
\label{subsubsec:ogd_name_model}
\begin{figure}[H]
    \centering
    \includegraphics[width=1\linewidth]{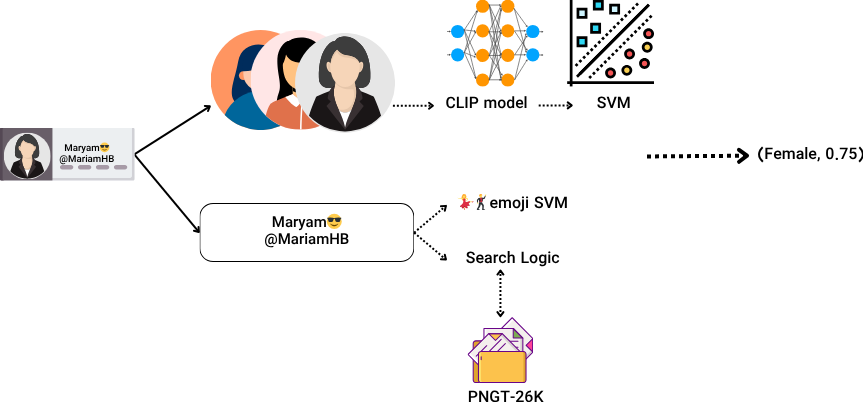}
    \caption{Design diagram for Open Gender Detection}
    \label{fig:design}
\end{figure}

The name-based inference module uses normalized Levenshtein distance\cite{nld} as defined in equation \ref{eq:normlized_levenshtein} where D(a,b) represents the standard Levenshtein distance, which is the minimum number of single-character edits (insertions, deletions, or substitutions) required to transform string "aa" into string "bb". The Levenshtein distance itself is computed recursively as shown in \ref{eq:edit_distance}. This distance metric is used along with PNGT-26K to search for names and outputs a probabilistic guess of their assigned gender by finding the top K (where K is an odd number to avoid ties) similar to the input. To improve the search pipeline and decrease cases of false mismatch whilst searching, the inputs are normalized as described in \ref{subsubsec:data_preprocessing}. To ensure extensibility and modularity, the implementation and design of this module allows the user to replace PNGT-26K with any other language, provided a name-gender dataset with the same format. This makes the system usable for a wider range of users. The overall architecture of the system is demonstrated in Figure \ref{fig:design}.

\begin{equation}
\label{eq:normlized_levenshtein}
d_{\text{lev}}(a, b) = \frac{D(a, b)}{\max(|a|, |b|)}
\end{equation}
\begin{equation}
\label{eq:edit_distance}
D(i, j) =
\begin{cases}
    \max(i, j), & \text{if } \min(i, j) = 0 \\
    \min
    \begin{cases}
        D(i-1, j) + 1, \\
        D(i, j-1) + 1, \\
        D(i-1, j-1) + 1_{[a_i \neq b_j]}
    \end{cases}, & \text{otherwise}
\end{cases}
\end{equation}

\subsubsection{Image-Based Inference Model}
\label{subsubsec:ogd_image_model}
Profile pictures in social media can be classified into two broad categories. These pictures might include or illustrate a human being in some shape, form, or style, or not include a human and contain writings, logos, landscapes, etc. Images in each class might contain clues about the user's gender. By utilizing the zero-shot classification abilities of CLIP\cite{radford2021learningtransferablevisualmodels}, textual prompts such as the ones shown in Table \ref{tab:clip_prompts} could be used to detect each of the mentioned classes. However, due to the inherent variety in the input data, despite this zero-shot classification ability, the diverse nature of profile pictures in social media necessitates training a classifier using a set of labeled pictures to increase generalization and avoid handcrafted prompts that can cause edge cases and missclassifications.
The method used for image-based gender classification is the classic technique of embedding vectors generated by an encoder model, clip-vit-base-patch32\footnote{\href{https://huggingface.co/openai/clip-vit-base-patch32}{https://huggingface.co/openai/clip-vit-base-patch32}} and then training or fine-tuning a classifier using labeled samples. After gathering a labeled dataset of 80K profile pictures of men and 80K profile pictures of women from social media, an SVM was trained on this dataset and was later used to classify embeddings from the encoder model. This module can be given a list of users' profile pictures as input, and will use a weighted mean approach to give the final result based on the results for each profile picture.

\begin{table}[h]
  \caption{Example textual prompts for zero-shot image classification using CLIP.}
  \centering
  \begin{tabular}{lp{9cm}}
    \toprule
    \textbf{Category} & \textbf{Example Prompts} \\
    \midrule
    Human vs. Non-human & 
    ``A photo of a person'', ``A portrait of a human face'', 
    ``A cartoon character'', ``A logo or symbol'', 
    ``A landscape photo'' \\
    \addlinespace
    Gender-related (Human) & 
    ``A photo of a man'', ``A photo of a woman'', 
    ``A boy's portrait'', ``A girl's portrait'' \\
    \addlinespace
    Non-human Subtypes & 
    ``A company logo'', ``A motivational quote image'', 
    ``A pet or animal photo'',  ``An abstract artwork'' \\
    \bottomrule
  \end{tabular}
  \label{tab:clip_prompts}
\end{table}

\subsection{Nominalist: An Agentic Framework for Username Generation}
\label{sec:nominalist}

\subsubsection{Multi-Agent System Architecture}
\label{subsubsec:nominalist_design}
\begin{figure}[H]
    \centering
    \includegraphics[width=1\linewidth]{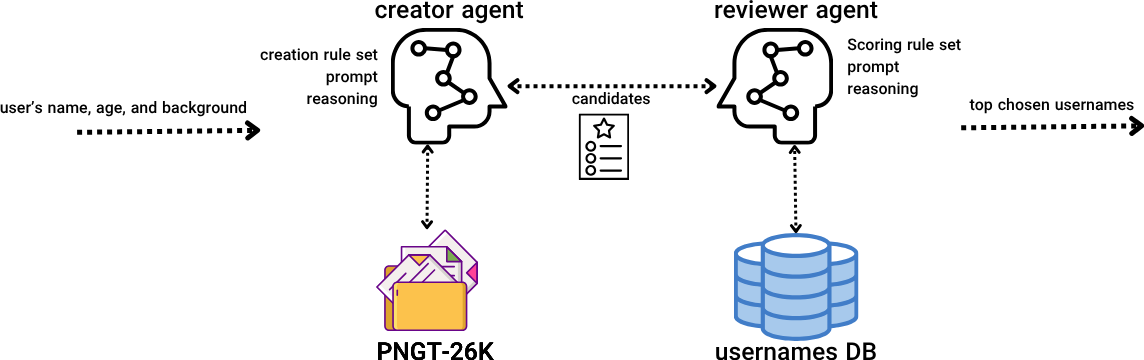}
    \caption{Design diagram for Nominalist}
    \label{fig:nominalist-design}
\end{figure}

Nominalist is a ready-to-use multi-agent framework that can utilize any LLM API, along with a name dataset to suggest usernames for a given user. As shown in Figure \ref{fig:nominalist-design}, the system is composed of two agents, a creator and a reviewer agent. When given the required inputs, the creator agent uses the PNGT-26K dataset to retrieve the English transliteration of the given name along with its gender. Afterwards, a predetermined set of rules is utilized to generate multiple variants of the input, which are considered valid usernames. Then, using the generative model API mentioned earlier, more candidates are created to ensure diversity and quality of the final results. The candidates created by the creator agent are then given as input to the reviewer agent, which uses an LLM, a set of rules, an existing usernames database, and heuristics to assess its input for uniqueness, validity, and desirability for the user. Finally, the K top usernames are chosen based on their score in this section and given as the final output of the system.

\subsubsection{Generative Core Structure} \label{subsubsec:nominalist_generative_core} The generative core of Nominalist
employs a hybrid approach combining rule-based transformations with large language model inference. The CreatorAgent
implements a dual-pathway generation mechanism that produces 10-12 username candidates through two distinct methods. The
first pathway utilizes nine predefined transformation rules, including underscore insertion, numerical suffixes, year appending, dot
notation, prefix/suffix addition, case normalization, space removal, and random suffix generation. These configurable rules ensure systematic and up-to-date coverage of common username patterns while maintaining deterministic behavior for consistent baseline generation.
The second pathway leverages any OpenAI-like API through a carefully crafted prompt engineering approach. The
system generates 5-6 creative username variants using a sampling temperature setting of 0.8 to encourage diversity while maintaining coherence. The prompt framework specifies strict constraints including character length limits (4-20 characters),
character set restrictions (alphanumeric and underscores only), and professional appropriateness criteria. The AI-generated candidates undergo post-processing validation using regular expressions to ensure compliance with platform-specific username requirements and remove any potentially problematic content. 

\section{Experiments}
\subsection{PNGT-26K characteristics}
\label{subsubsec:dataset_characteristics}
\paragraph{Male-female distribution}
The PNGT-26K dataset, after undergoing all the post-processing steps documented in \ref{sec:dataset_creation}, was analyzed for gender distribution and was found to consist of 65\% and 35\% male and female names, respectively. This imbalance was found to be partly due to the high number of male-specific names that are created as a combination of other male-specific names. For instance, the name «\FR{محمد}» can be fused to other Persian or  Arabic names to create new names, which are valid and even widely used, like «\FR{محمد‌علی}» or «\FR{محمد‌رضا}», which are among some of the most used names in Iran as a Persian-speaking country.

\begin{figure}[htp]
    \centering
    \subfigure[Histogram of Persian Name Lengths]{
        \includegraphics[width=0.48\textwidth]{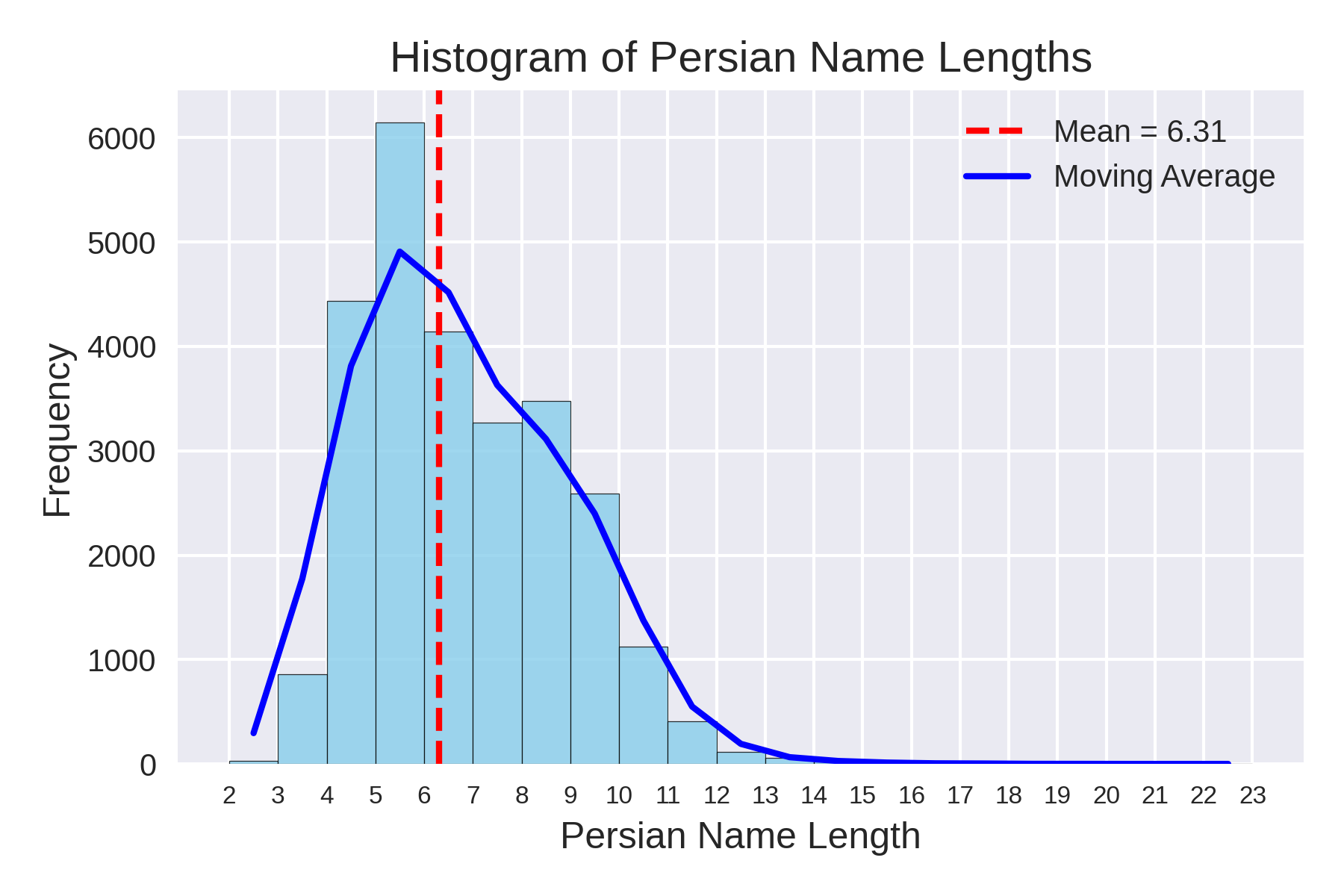}
        \label{fig:persian_names}
    }
    \hfill
    \subfigure[Histogram of English Name Lengths]{
        \includegraphics[width=0.48\textwidth]{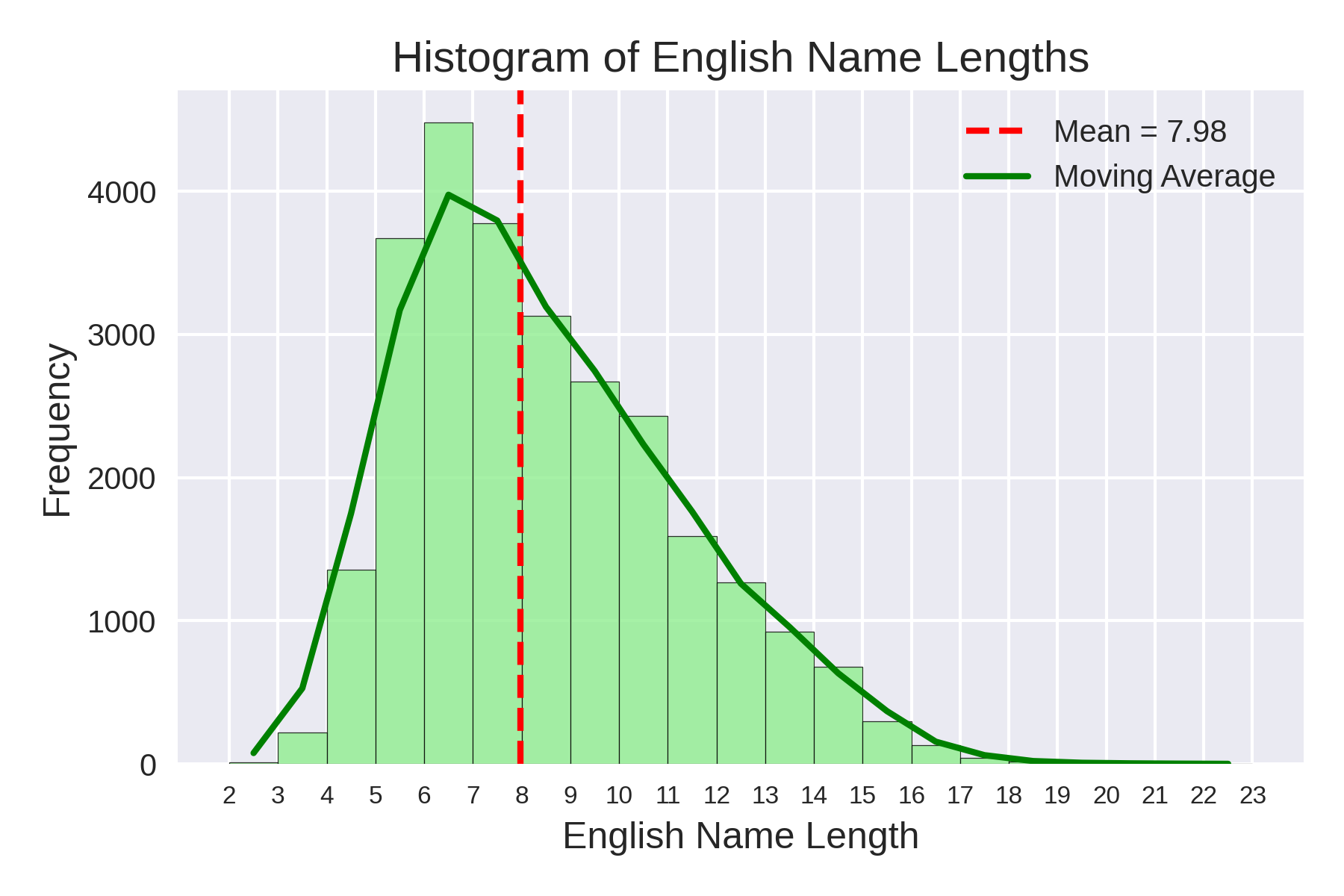}
        \label{fig:english_names}
    }
    \caption{Comparison of Name Length Distributions}
    \label{fig:name_distributions}
\end{figure}

\paragraph{Distribution of Name Lengths in Persian and English}
The length feature of both Persian names and their transliterations followed a chi-squared ($\chi^2$) with the mean residing at 6.30 and 7.97, respectively. The distribution histogram of the English and Persian names is visualized in \ref{fig:english_names} and \ref{fig:persian_names}, respectively.

\paragraph{Character Distribution by Gender}
Prior studies, such as \cite{he2020long}, have thoroughly documented orthographic patterns between male and female names. Tables \ref{tab:char_frequency_male} and \ref{tab:char_frequency_female} report the top 5 most-used characters in Persian names. The occurrence of each Persian character was calculated separately for male and female names, and several notable trends are visible in Tables \ref{tab:char_frequency_male} and \ref{tab:char_frequency_female}. These patterns reflect the inherent differences in naming conventions between genders and provide an opportunity for further analysis of cultural and linguistic differences in Persian naming practices.

\begin{table}[ht]
\centering
\caption{Top 5 Most Repeated Characters for Males ordered by count}
\label{tab:char_frequency_male}
\begin{tabular}{cccccc}
\toprule
Gender & Persian Count & Persian Character  & English Character & English Count \\
\midrule
\multirow{5}{*}{male} & 15475 & \FR{ا} & a & 30805 \\
      & 12076 & \FR{ی} & h & 11889 \\
      & 9288 & \FR{م}  & m & 10490 \\
      & 9118 & \FR{ر}  & i & 9989  \\
      & 7055 & \FR{د}  & r & 9136  \\
\bottomrule
\end{tabular}
\end{table}
\begin{table}[ht]
\centering
\caption{Top 5 Most Repeated Characters for Females ordered by count}
\label{tab:char_frequency_female}
\begin{tabular}{cccccc}
\toprule
Gender & Persian Count & Persian Character  & English Character & English Count \\
\midrule
\multirow{5}{*}{female} & 8204 & \FR{ا}  & a & 14072 \\
      & 5481 & \FR{ی}  & h & 6909  \\
      & 4773 & \FR{ن}  & n & 4791  \\
      & 4418 & \FR{ر}  & i & 4637  \\
      & 4026 & \FR{ه}  & r & 4459  \\
\bottomrule
\end{tabular}
\end{table}

\subsection{Probability Fusion in Open Gender Detection}
\label{subsubsec:ogd_fusion}
For fusing the output of the two modules of this framework, a combination of a voting mechanism and weighted mean was used. The proposed method first checks for the result of the name-based inference, and if the confidence of the output towards male or female is larger than a user-defined threshold, the output will match the gender and probability given by the name-based inference. This choice was made due to the comprehensiveness of the underlying lookup, which reduces the need to use the resource-consuming method of image-based inference. In the case where the name-based inference lacks enough confidence, the output of the image-based inference is used as a second judgment. In this scenario, if the outputs of the two models share the same gender, the final result would also be the same; otherwise, a tie occurs, and the proposed method for tie-breaking is to use a weighted sum of the output probabilities of the two models. This method enables the system to be easily configured for various environments by letting users use any desired weights for the weighted sum operation.

\subsection{Multi-Agent implementation in Nominalist} \label{subsubsec:nominalist_interaction} The multi-agent interaction protocol follows a sequential pipeline architecture with well-defined interfaces between components. The CreatorAgent initiates the workflow by receiving an input name and interfacing with the NameService to obtain English transliterations from the PNGT-26K or any other dataset with the same format. The agent then orchestrates the parallel execution of rule-based and AI-powered generation pathways, collecting and
deduplicating results to ensure exactly 10-12 unique candidates. 
The ReviewerAgent subsequently processes these candidates through a three-stage evaluation pipeline. First, it checks against existing username databases, filtering out unavailable
options. Second, it employs the same OpenAI-like API with a specialized ranking prompt that evaluates candidates across five dimensions:
memorability, professional appearance, typing ease, uniqueness, and overall appeal. The AI evaluation uses a lower temperature
setting (0.3) to ensure consistency. Third, traditional heuristic scoring is applied based on username length
optimization (6-15 characters preferred), numerical content analysis, and pattern matching for readability. 
The final ranking employs a weighted combination algorithm that merges AI-derived scores (60\% weight) with traditional heuristic scores (40\% weight), providing robustness against AI model failures while leveraging advanced semantic understanding.

This flexible design makes the mentioned framework easily usable for any given source language and easily integrable into other software. When using languages other than Persian, the only required change is replacing PNGT-26k with a similarly formatted dataset based on the source language requirements. As for integrating into existing backend systems, the software is designed to require minimal change, only in the case of database access.

\subsubsection{Nominalist Technical Details} \label{subsubsec:nominalist_implementation} Nominalist is designed as a containerized, ready-to-use framework distributed through Docker for seamless deployment and cross-platform compatibility. The system supports any OpenAI-compatible API through configurable base URL endpoints, making it particularly suitable for local LLM deployments where the relatively low complexity of username generation tasks can be efficiently handled by smaller, locally-hosted models without requiring large-scale cloud infrastructure.
The framework's language adaptability is achieved through its modular dataset interface, where the PNGT-26K dataset can be substituted with any name translation dataset following the same format structure. This design enables extension to multiple languages and cultural naming conventions beyond the current English transliteration scope, allowing practitioners to adapt the system to region-specific username generation requirements.
The architecture prioritizes extensibility through well-defined service interfaces and multi-agent design. Additional generation rules, scoring mechanisms, or validation criteria can be integrated through the existing agent interfaces without requiring modifications to the core system. The framework supports expansion with new agents, alternative ranking algorithms, or specialized domain-specific username requirements through its plugin-compatible structure.

\section{Conclusion}
This paper addresses critical gaps in Persian name processing and digital identity generation through the introduction of three interconnected contributions. The \textit{PNGT-26K} dataset provides the first large-scale, publicly available resource for Persian names with systematic English transliterations, comprising 26,000 name entries, offering researchers and practitioners a foundation for developing culturally-aware NLP applications. 
The \textit{Open Gender Detection} framework demonstrates the practical application of our dataset through a multimodal approach that combines name-based inference using normalized Levenshtein distance with image-based classification trained on 160K profile pictures to ensure generalization. The system's probabilistic output and fusion methodology provide flexibility for various scenarios while maintaining transparency about prediction confidence. The modular design allows for adaptation to other languages by substituting the underlying name dataset, enhancing the framework's global applicability.

The \textit{Nominalist} framework introduces a novel agentic AI approach to username generation that moves beyond simple heuristics to provide culturally aware, personalized digital identity suggestions. The multi-agent architecture combining rule-based transformations with large language model creativity offers both deterministic baseline generation and diverse creative options. The framework's containerized design and API compatibility facilitate integration into existing systems.

Future work should focus on expanding the dataset to include regional variations within Persian-speaking communities and developing more sophisticated transliteration standardization methods. Additionally, exploring the integration of semantic gender cues alongside morphological patterns could further improve gender detection accuracy. The agentic framework design opens opportunities for specialized domain-specific username generation and multi-language adaptation.

\bibliographystyle{unsrt}
\bibliography{references}

\begin{thebibliography}{10}

\bibitem{vanhelene2024inferring}
Alexander~D. VanHelene, Inder Khatri, C.~Beau Hilton, Smrati Mishra, E.~Deniz Gamsiz~Uzun, and Jeremy~L. Warner.
\newblock Inferring gender from first names: Comparing the accuracy of genderize, gender api, and the gender r package on authors of diverse nationality.
\newblock {\em PLOS Digital Health}, 3(10):e0000456, 2024.

\bibitem{babelstreet2022persian}
{Babel Street}.
\newblock What's in a persian name?
\newblock Babel Street Blog, 2022.
\newblock Accessed: July 10, 2025.

\bibitem{kazerani2018examining}
Maryam Kazerani and Maryam Shekofteh.
\newblock Examining the problems and inconsistencies in the interpolation of english transliterated names of persian language researchers in citation databases.
\newblock 2018.

\bibitem{ssa_baby_names}
{U.S. Social Security Administration}.
\newblock Beyond the top 1000 names.
\newblock \url{https://www.ssa.gov/oact/babynames/limits.html}, 2023.
\newblock Accessed: July 10, 2025.

\bibitem{mullen2014gender}
Lincoln Mullen.
\newblock {\em gender: Predict Gender from Names Using Historical Data}, 2021.
\newblock R package version 0.6.0.

\bibitem{hu2021name}
Yifan Hu, Changwei Hu, Thanh Tran, Tejaswi Kasturi, Elizabeth Joseph, and Matt Gillingham.
\newblock {What's in a name? -- gender classification of names with character based machine learning models}.
\newblock {\em Data Mining and Knowledge Discovery}, 35(4):1537--1563, 2021.

\bibitem{perez2017learning}
Carlos P{\'e}rez~Estruch, Roberto Paredes~Palacios, and Paolo Rosso.
\newblock Learning multimodal gender profile using neural networks.
\newblock In Ruslan Mitkov and Galia Angelova, editors, {\em Proceedings of the International Conference Recent Advances in Natural Language Processing, {RANLP} 2017}, pages 577--582, Varna, Bulgaria, September 2017. INCOMA Ltd.

\bibitem{5206848}
Jia Deng, Wei Dong, Richard Socher, Li-Jia Li, Kai Li, and Li~Fei-Fei.
\newblock Imagenet: A large-scale hierarchical image database.
\newblock In {\em 2009 IEEE Conference on Computer Vision and Pattern Recognition}, June 2009.

\bibitem{blei2003latent}
David~M Blei, Andrew~Y Ng, and Michael~I Jordan.
\newblock Latent dirichlet allocation.
\newblock {\em Journal of machine Learning research}, 3(Jan):993--1022, 2003.

\bibitem{pennebaker2001linguistic}
James~W Pennebaker, Martha~E Francis, Roger~J Booth, et~al.
\newblock Linguistic inquiry and word count: Liwc 2001.
\newblock {\em Mahway: Lawrence Erlbaum Associates}, 71(2001):2001, 2001.

\bibitem{genderize_api}
{Gender API}.
\newblock Gender api - determines the gender of a first name.
\newblock \url{https://gender-api.com/en/api-docs}, 2024.
\newblock Accessed: July 10, 2025.

\bibitem{abdulmageed2020arbert}
Muhammad Abdul-Mageed, Amir El-Gohary, and Chiyu Kulkarni.
\newblock {AraBERT: Transformer-based Model for Arabic Language Understanding}.
\newblock {\em arXiv preprint arXiv:2003.00104}, 2020.

\bibitem{hassanzadeh2022effects}
Mahsa Kaveh, Mahdieh Mirzabeigi, Hajar Sotudeh, and Amirsaeid Moloodi.
\newblock The effects of the challenges in the transliteration of persian names into english on the recall of retrieved results in the web of science.
\newblock {\em Scientometrics}, 127(2):1099–1128, January 2022.

\bibitem{cjki2023persian}
{The CJK Dictionary Institute, Inc.}
\newblock Database of persian names.
\newblock CJKI Data Products, 2023.
\newblock Accessed: July 10, 2025.

\bibitem{titanz123_kaggle}
titanz123.
\newblock Persian names with gender and transliteration data.
\newblock Kaggle, 2021.
\newblock Accessed: July 10, 2025.

\bibitem{zeoses_github}
zeoses.
\newblock persian-gender-detection.
\newblock GitHub repository, 2022.
\newblock Accessed: July 10, 2025.

\bibitem{armanyazdi_github}
armanyazdi.
\newblock persian-gender-detection-go.
\newblock GitHub repository, 2022.
\newblock Accessed: July 10, 2025.

\bibitem{peymanslh_npm}
peymanslh.
\newblock persian-gender-detection.
\newblock NPM package, 2022.
\newblock Accessed: July 10, 2025.

\bibitem{peris2016identifying}
Yubin Wang, Tingwen Liu, Qingfeng Tan, Jinqiao Shi, and Li~Guo.
\newblock Identifying users across different sites using usernames.
\newblock {\em Procedia Computer Science}, 80:376--385, 2016.
\newblock International Conference on Computational Science 2016, ICCS 2016, 6-8 June 2016, San Diego, California, USA.

\bibitem{topolinski2017username}
Rita Silva, Nina Chrobot, Eryn Newman, Norbert Schwarz, and Sascha Topolinski.
\newblock Make it short and easy: Username complexity determines trustworthiness above and beyond objective reputation.
\newblock {\em Frontiers in Psychology}, 8, 12 2017.

\bibitem{sutskever2011generating}
Ilya Sutskever, James Martens, and Geoffrey Hinton.
\newblock Generating text with recurrent neural networks.
\newblock In {\em Proceedings of the 28th International Conference on International Conference on Machine Learning}, ICML'11, page 1017–1024, Madison, WI, USA, 2011. Omnipress.

\bibitem{rahalkar2020name}
Chaitanya Rahalkar.
\newblock Name-generator-rnn: A name generator implemented using a recurrent neural network from scratch.
\newblock GitHub repository, 2020.
\newblock Accessed: July 10, 2025.

\bibitem{sutskever2014sequence}
Ilya Sutskever, Oriol Vinyals, and Quoc~V. Le.
\newblock Sequence to sequence learning with neural networks, 2014.

\bibitem{iranianNamesDB}
Nikahd99.
\newblock Iranian names database by gender.
\newblock \url{https://github.com/nikahd99/iranian-Names-Database-By-Gender}.
\newblock Accessed: 2025-08-15.

\bibitem{persianNames}
titanz123.
\newblock Persian names.
\newblock \url{https://www.kaggle.com/datasets/titanz123/persian-names}.
\newblock Accessed: 2025-08-15.

\bibitem{persianNamesGender}
misssahar75.
\newblock Persian names gender.
\newblock \url{https://www.kaggle.com/datasets/misssahar75/persian-names-gender}.
\newblock Accessed: 2025-08-15.

\bibitem{doctor2024graphemicnormalizationpersoarabicscript}
Raiomond Doctor, Alexander Gutkin, Cibu Johny, Brian Roark, and Richard Sproat.
\newblock Graphemic normalization of the perso-arabic script, 2024.

\bibitem{hazm}
Roshan Research.
\newblock Hazm: Python library for persian nlp.
\newblock \url{https://github.com/roshan-research/hazm}.
\newblock Accessed: 2025-08-15.

\bibitem{deepseekai2025deepseekr1incentivizingreasoningcapability}
DeepSeek-AI, Daya Guo, Dejian Yang, Haowei Zhang, Junxiao Song, Ruoyu Zhang, Runxin Xu, Qihao Zhu, Shirong Ma, Peiyi Wang, Xiao Bi, Xiaokang Zhang, Xingkai Yu, Yu~Wu, Z.~F. Wu, Zhibin Gou, Zhihong Shao, Zhuoshu Li, Ziyi Gao, Aixin Liu, Bing Xue, Bingxuan Wang, Bochao Wu, Bei Feng, Chengda Lu, Chenggang Zhao, Chengqi Deng, Chenyu Zhang, Chong Ruan, Damai Dai, Deli Chen, Dongjie Ji, Erhang Li, Fangyun Lin, Fucong Dai, Fuli Luo, Guangbo Hao, Guanting Chen, Guowei Li, H.~Zhang, Han Bao, Hanwei Xu, Haocheng Wang, Honghui Ding, Huajian Xin, Huazuo Gao, Hui Qu, Hui Li, Jianzhong Guo, Jiashi Li, Jiawei Wang, Jingchang Chen, Jingyang Yuan, Junjie Qiu, Junlong Li, J.~L. Cai, Jiaqi Ni, Jian Liang, Jin Chen, Kai Dong, Kai Hu, Kaige Gao, Kang Guan, Kexin Huang, Kuai Yu, Lean Wang, Lecong Zhang, Liang Zhao, Litong Wang, Liyue Zhang, Lei Xu, Leyi Xia, Mingchuan Zhang, Minghua Zhang, Minghui Tang, Meng Li, Miaojun Wang, Mingming Li, Ning Tian, Panpan Huang, Peng Zhang, Qiancheng Wang, Qinyu Chen, Qiushi Du, Ruiqi Ge, Ruisong
  Zhang, Ruizhe Pan, Runji Wang, R.~J. Chen, R.~L. Jin, Ruyi Chen, Shanghao Lu, Shangyan Zhou, Shanhuang Chen, Shengfeng Ye, Shiyu Wang, Shuiping Yu, Shunfeng Zhou, Shuting Pan, S.~S. Li, Shuang Zhou, Shaoqing Wu, Shengfeng Ye, Tao Yun, Tian Pei, Tianyu Sun, T.~Wang, Wangding Zeng, Wanjia Zhao, Wen Liu, Wenfeng Liang, Wenjun Gao, Wenqin Yu, Wentao Zhang, W.~L. Xiao, Wei An, Xiaodong Liu, Xiaohan Wang, Xiaokang Chen, Xiaotao Nie, Xin Cheng, Xin Liu, Xin Xie, Xingchao Liu, Xinyu Yang, Xinyuan Li, Xuecheng Su, Xuheng Lin, X.~Q. Li, Xiangyue Jin, Xiaojin Shen, Xiaosha Chen, Xiaowen Sun, Xiaoxiang Wang, Xinnan Song, Xinyi Zhou, Xianzu Wang, Xinxia Shan, Y.~K. Li, Y.~Q. Wang, Y.~X. Wei, Yang Zhang, Yanhong Xu, Yao Li, Yao Zhao, Yaofeng Sun, Yaohui Wang, Yi~Yu, Yichao Zhang, Yifan Shi, Yiliang Xiong, Ying He, Yishi Piao, Yisong Wang, Yixuan Tan, Yiyang Ma, Yiyuan Liu, Yongqiang Guo, Yuan Ou, Yuduan Wang, Yue Gong, Yuheng Zou, Yujia He, Yunfan Xiong, Yuxiang Luo, Yuxiang You, Yuxuan Liu, Yuyang Zhou, Y.~X. Zhu,
  Yanhong Xu, Yanping Huang, Yaohui Li, Yi~Zheng, Yuchen Zhu, Yunxian Ma, Ying Tang, Yukun Zha, Yuting Yan, Z.~Z. Ren, Zehui Ren, Zhangli Sha, Zhe Fu, Zhean Xu, Zhenda Xie, Zhengyan Zhang, Zhewen Hao, Zhicheng Ma, Zhigang Yan, Zhiyu Wu, Zihui Gu, Zijia Zhu, Zijun Liu, Zilin Li, Ziwei Xie, Ziyang Song, Zizheng Pan, Zhen Huang, Zhipeng Xu, Zhongyu Zhang, and Zhen Zhang.
\newblock Deepseek-r1: Incentivizing reasoning capability in llms via reinforcement learning, 2025.

\bibitem{rahman2024llms}
Samira Rahman and Md~Jabir Rahman.
\newblock Llms for gender prediction: A comparative study.
\newblock 2024.

\bibitem{radford2021learningtransferablevisualmodels}
Alec Radford, Jong~Wook Kim, Chris Hallacy, Aditya Ramesh, Gabriel Goh, Sandhini Agarwal, Girish Sastry, Amanda Askell, Pamela Mishkin, Jack Clark, Gretchen Krueger, and Ilya Sutskever.
\newblock Learning transferable visual models from natural language supervision, 2021.

\bibitem{nld}
Li~Yujian and Liu Bo.
\newblock A normalized levenshtein distance metric.
\newblock {\em IEEE Transactions on Pattern Analysis and Machine Intelligence}, 29(6):1091--1095, 2007.

\bibitem{he2020long}
Katherine He.
\newblock Long-term sociolinguistics trends and phonological patterns of american names.
\newblock {\em Proceedings of the Linguistic Society of America}, 5(1):616--622, 2020.

\end{thebibliography}

\end{document}